\documentclass[a4paper,twoside]{article}

\usepackage{epsfig}
\usepackage{subcaption}
\usepackage{calc}
\usepackage{amssymb}
\usepackage{amstext}
\usepackage{amsmath}
\usepackage{amsthm}
\usepackage{multicol}
\usepackage{pslatex}
\usepackage{apalike}

\usepackage{subcaption}
\usepackage{booktabs} % for professional tables
\usepackage{comment}
\usepackage{tabularx}
\usepackage{threeparttable}
\usepackage{multirow}
\usepackage{hyperref}

\usepackage{algorithm}
\usepackage{algorithmic}
\usepackage{etoolbox}% http://ctan.org/pkg/etoolbox
\AtBeginEnvironment{algorithmic}{\let\textbf\relax}
\usepackage{SCITEPRESS}     % Please add other packages that you may need BEFORE the SCITEPRESS.sty package.

\begin{document}
%find new title
\title{Multi-Agent Transfer Learning in Reinforcement Learning-Based Ride-Sharing Systems}
%\author{\authorname{Alberto Castagna \sup{1}, Ivana Dusparic\orcidAuthor{0000-0003-0621-5400} \sup{1}}
\author{\authorname{Alberto Castagna and Ivana Dusparic}
\affiliation{School of Computer Science and Statistics, Trinity College Dublin, Ireland}
%\affiliation{\sup{2}Department of Computing, Main University, MySecondTown, MyCountry}
\email{\{acastagn, ivana.dusparic\}@tcd.ie}
}


%TODO UP TO 200. WE'VE 237
\abstract{
Reinforcement learning (RL) has been used in a range of simulated real-world tasks, e.g., sensor coordination, traffic light control, and on-demand mobility services. However, real world deployments are rare, as RL struggles with dynamic nature of real world environments, requiring time for learning a task and adapting to changes in the environment.
Transfer Learning (TL) can help lower these adaptation times. In particular, there is a significant potential of applying TL in multi-agent RL systems, where multiple agents can share knowledge with each other, as well as with new agents that join the system. To obtain the most from inter-agent transfer, transfer roles (i.e., determining which agents act as sources and which as targets), as well as relevant transfer content parameters (e.g., transfer size) should be selected dynamically in each particular situation. As a first step towards fully dynamic transfers, in this paper we investigate the impact of TL transfer parameters with fixed source and target roles. Specifically, we label every agent-environment interaction with agent's epistemic confidence, and we filter the shared examples using varying threshold levels and sample sizes. We investigate impact of these parameters in two scenarios, a standard predator-prey RL benchmark and a simulation of a ride-sharing system with 200 vehicle agents and 10,000 ride-requests.
}

\onecolumn \maketitle \normalsize \setcounter{footnote}{0} \vfill

\section{INTRODUCTION}
\label{sec:intro}

Reinforcement learning (RL) has shown good performance in addressing a variety of tasks, from naive games to more complicated problems that require synchronization.
RL enables an intelligent agent optimize its performance in a specific task, however, when a change in the underlying environment or a task occurs, the performance of an agent often sharply decrease. Thus, RL often performs very well in a simulated environment but struggles in real world evolving environments since it requires a constant updating of the knowledge.

To discuss issues of applying RL in such real world applications, in this work we apply multi-agent RL for the ride-request assignment in a Mobility On-Demand (MoD) system. Ride-request assignment is a widely studied task, addressed as global optimization problem~\cite{Fagnant2017,Wen2017,yang2019} or locally by distributing the control at vehicles level~\cite{castagna2020demand,castagna2020d2r2,samod2020,alabbasi2019deeppool}. Recent works adopt the use of RL where an agent controls a single vehicle, as we do in this paper. The type of changes that RL might need to adapt to in an MoD system, but which have not been addressed in literature, can, for example, be that road layouts change, e.g., due to closure, or that demand load or pattern changes over time, temporarily or permanently. 

A possible solution to these issues is to apply transfer learning (TL) in conjunction with RL. Once encounters a new situation, an agent can reuse knowledge from other agents or its previous collected experiences in order to boost its initial performance in the new situation. For example, MoD system in \cite{castagna2020d2r2} replicates full knowledge from one agent to others. When a new agent joins the fleet, it can inherit knowledge good enough to be productive with no delay. Another approach to transferring knowledge across similarly defined tasks is by fine tuning previous performance; this process replicates the knowledge from one agent to the other and afterwards, the receiving agent performs a refinement step to tune the received knowledge in its own settings and environment. 
As an example, \cite{wang2018DRLknowledgetransfer} proposes fine-tuning in rider-driver assignment problem across scenarios with different underlying dynamics, e.g.,  demand patterns and road structure. However, fine-tuning protocols are usually applied when the differences are relatively small. The time required for the receiving agent to adapt the received information is related to the gap magnitude between source and destination conditions. As this become wider, as more time the agent needs to refine the knowledge.

Transferring from a source to a very different target task, could lead to so-called negative transfer, as result the second agent requires longer time to solve the task compared to an agent that is learning from scratch. When negative transfer happens, on top of the time needed to solve the task, receiving agent need additional time to forget the wrong injected knowledge. Therefore, while TL can significantly improve the performance of an RL system, the knowledge transferred needs to be carefully selected and integrated, to prevent negative transfer resulting in even worse performance than without deploying TL.

TL can also be applicable when multiple agents are learning a task simultaneously. By exploiting transfer techniques, agents can collaborate to achieve a faster convergence rate. When applied in real-time, further challenges need to be addressed, such as selecting the best agent as source of transfer among those available. 

As the first step towards this vision in which agents in a multi-agent system continually share the knowledge with evolving parameters, in this paper we study the impact of parameters used when selecting the knowledge to be transferred, but in an offline transfer scenario, where the roles of sender and receiver are well defined.
Specifically, we bootstrap the agent with agent-environment interactions collected from other agents across similar scenarios. Transferred knowledge is annotated with a value depicting agent's confidence in that particular experience, as in~\cite{ilhan2019teaching,Silva2020UncertaintyAwareAA}.
We study TL performance using different batches of experiences, varying quantity of information shared as well as its "quality", as determined by adjusting the confidence threshold level.

We perform the transfer across agents performing the same task, but in completely different underlying environment dynamics. The target agent has to perform a full training cycle in the novel environment, but prior to that it inherits knowledge from another trained agent in order to bootstrap the performance achieved.%might 

%Thanks to the form of knowledge transferred and to the use of an external network in estimating confidence, this technique can be applied to a variety of RL techniques, ranging from tabular methods to deep reinforcement learning algorithms. Furthermore, an agent with a tabular representation could transfer to an agent equipped with neural network and vice versa.
%ok but what we do?
%In this work, we analyse the impact of reusing previous collected experience 
% agent-environment interactions %guess to specific for now.
%from a source to a target agent, where both agents are equally defined and are addressing the same task but with different environment dynamics.
%Compared to fine-tuning, our target agent has to perform a full training cycle on the novel task, but prior to that it inherits knowledge from another trained agent in order to boost the performance achieved.%might seem like we are doing more than fine-tuning but our end goal is to go on online tl
%before beginning this phase it inherit knowledge from another trained agent that has already addressed the same task 

In our initial investigations presented in this paper, we focus on a simple case where just two agents are involved and the underlying learning model is represented by a neural network for both. Transfer goal focuses on improving the performance on receiving side and therefore, the communication is one way. %More complicated scenarios with multiple agents can be addressed in the same way.

For our implementation, we combine Proximal Policy Optimization (PPO)~\cite{PPOagentOpenAI}, with external knowledge in form of agent-environment interactions. Unlike off-policy algorithms that use two different policies to sample and optimize, on-policy PPO assumes that the experience used to optimize the policy is drawn by the very same policy. Therefore, we extend PPO to bootstrap external information to perform few steps of PPO gradient descent optimization before exploration begins.
%However, in our experiments we observed an improvement on the performance when external information are processed at the very begin before exploiting the agent's policy.
%We further discuss this topic in Sec.~\ref{sec:fwork}.

Evaluation is performed in two application areas: benchmark environment predator-prey and ride sharing simulation. The latter experiments are carried out on a simulated MoD environment with ride-sharing enabled vehicles. Requests are obtained from New York City dataset~\cite{NYC_data} and simulation performed in the Simulator of Urban MObility (SUMO). Finally, we compare the performance against baselines: (1) policy transfer and (2) learning from scratch.

Contribution of this paper is therefore twofold: 1) we evaluate the impact of transferring confidence labelled agent-environment interactions by varying the transfer batch size and filtering using different threshold level; 2) we study applicability of TL in a Mobility-on-Demand system with ride-sharing~(RS) enabled vehicles under different demand patterns. %3) we extend an on-policy RL algorithm to bootstrap external knowledge.

The rest of this paper is organised as follows.  Section~\ref{sec:background} reviews relevant work related to experience reusing in transfer learning applied to RL. Section~\ref{sec:proposal} describes in detail the process used to share experience within this research, Section~\ref{sec:simulation} presents the scenarios, Section~\ref{sec:results} discusses the results and discuss the findings while Section~\ref{sec:fwork} concludes this manuscript by giving further research directions.
\section{BACKGROUND}\label{sec:background}
This section introduces the reader to previous related work on transfer learning and on the rider-driver matching problem for mobility on-demand system, finally discusses available methodologies to estimate agent's confidence.

\subsection{Rider-Driver Matching Problem}

%RS ADDRESSED BY RL
Rider-driver matching problem is addressed using a wide range of both centralized and decentralized techniques, with different inputs and assumptions. Generally, centralized aims to directly optimize the global performance, i.e., minimising the travelled distance~\cite{yang2019,Fagnant2017,Wen2017} while decentralized addresses the problem from a single vehicle perspective where each vehicle is independent and unaware of others doing.
Decentralized approaches often rely on RL where each vehicle is controlled by a single agent~\cite{castagna2020demand,castagna2020d2r2,gueriau2018samod,samod2020,alabbasi2019deeppool}.
%In this paper, we replicate a decentralized version from ~\cite{castagna2020d2r2}, where we disable the rebalancing scheduler to avoid external influences over the performance and replace the underlying engine with a widely adopted simulator of urban mobility, SUMO~\cite{sumo2018}.

\textbf{\title{Transfer Learning in rider-driver Assignment}}

%TRANSFER LEARNING IN RIDER-DRIVER ASSIGNMENT
Within rider-driver dispatch problem there are not many works that bootstrap external knowledge applying TL.
Previously, \cite{castagna2020d2r2} designed a multi-agent collaborative algorithm to obtain a main knowledge that is then replicated to other agents. In detail, multiple agents were learning to satisfy ride-requests in a iterative way where the set of available requests was composed by unserved requests from previous vehicles.
Finally, obtained knowledge is replicated to other agents and no further refinement steps are performed.

On the other hand,  ~\cite{wang2018smartcity,wang2018DRLknowledgetransfer}, bootstrap previous acquired knowledge with a fine-tuning process to leverage a pre-trained learning model.
Here, authors use a deep learning model and transfer a pre-trained neural network to target environments with different demand pattern and road infrastructure.
This technique limits the range of applicable algorithms to those with a neural network underlying as it is feasible to transfer just the layers weights, i.e., is not possible to transfer knowledge to tabular model. Furthermore, any change in the neural network architecture require further steps to adapt the pre-trained layers.

\cite{wan2021pattern} proposes an approach to transfer from source to a target environment. The goal is to fine-tune knowledge by adapting the source value-function to a target task with a concordance penalty that expresses the similar patterns between the two tasks, i.e. demand pattern. Thus, it can be applied to a range of RL techniques regardless the underlying representation. Nevertheless, a strong limitation of this work is the need of prior knowledge over the relation between source and target tasks.

\subsection{Transfer Learning}

Transfer learning is a widely used technique in a variety of fields, \cite{zhuang2020comprehensive} provides an overview of the state of the art approaches across many domains. In this work we apply TL from agent to agent. We analyse the two agents case and from now on, we refer to the agent that is providing information as source agent, while the other as target agent. The latter processes external knowledge to boost the performance within its task.

%When applied to reinforcement learning, we can distinguish two categories, task-to-task and agent-to-agent transfer. For sake of simplicity we discuss the simpler case where two agents are involved, from now one we refer to the agent that is providing informations as  source agent, while the other is called target agent and exploits external knowledge.
When transferring across agents, tasks are normally equally defined, therefore state representations, action sets and reward models are the same.
Despite these similarities, there can be some intrinsic characteristics of the environment that may differ, e.g., within MoD scenario, different configurations of the road network, different traffic loads or different demand pattern.

Existing applications of TL applied to RL are many, varying by type of transferred knowledge and technique used. For example, \textit{experience sharing} where agents share experience in form of agent-environment interactions~\cite{cruz2019pretraining,lazaric2008transfersamples} as we do within this work,
\textit{policy transfer} where a trained model is replicated to a novel agent as in~\cite{wang2018DRLknowledgetransfer}, \textit{advice-based} where an expert or another agent is available on-demand to support the learning phase of an agent~\cite{Fachantidis_2017,Silva2020Uncertainty-aware}, and finally by more sophisticated and task-related techniques as in~\cite{wan2021pattern}. Previous work also investigated the impact of transfer parameter selection in the context of tabular RL ~\cite{Taylor2019ptl}, and concluded that frequency and size of the transfer, as well as how is the knowledge incorporated on the target agent, has significant impact on the performance of the target agents.

%Existing transfer learning application to reinforcement learning problem, could be classified by the techniques used, \textit{experience reusing} where an agent reuses part of previous collected experience, \textit{transfer of policy}, where a part or a whole policy is replicated, \textit{learning from demonstration}, where an agent learns from example provided by an expert entity, \textit{reward shaping}, where the agent learning is driven by the changing within the reward function, \textit{inter-task mapping}, where an agent learns to address a variety of tasks simultaneously, and finally \textit{advice-based}, where an agent can rely on another entity to ask for advices when needed.

%\subsection{Proximal Policy Optimization}

%In respect of the scenarios analysied within this research, the experiments carried out on predator-prey environment~\ref{fig:pp} apply an agent-to-agent transfer while those on mobility on-demand system applies task to task, since the dyanmics are different. 
\begin{comment}
epsitemic confidence level of the agent the RND work + multi-heads nn

reusing experience from one to other with other polocies, i.e., latest episodes...
\end{comment}
\subsection{Confidence Estimator}

For estimating the agent's confidence within a state are available several possibilities that suit tabular and deep learning models.
A naive approach could be to count the agent visits over states or to  define a function over it as in~\cite{zhu2020learning}. 
However, this becomes unfeasible whenever the state space is too wide or continuous as in our MoD scenario.
More sophisticated models range from the use of a Gaussian model with a confidence function defined over it~\cite{Taylor2018improvingRL}, changes on neural network structure~\cite{Silva2020Uncertainty-aware} and finally using external tool to compute it as Random Network Distillation (RND)~\cite{burda2018RND}. RND has been presented as a tool to regulate exploration by curiosity but has been already used as uncertainty estimator for a RL agent by~\cite{ilhan2019teaching}. A neural network is trained to predict the output of a target rough network, while uncertainty is defined as discrepancy across the two networks output.

\section{CONFIDENCE-BASED TRANSFER LEARNING}
\label{sec:proposal}
Within this section we introduce the design of transfer learning approach we have implemented to evaluate confidence and sample size impact in transfer learning. %This paper stands as introductory study on bootstrapping external knowledge across multiple agents, however, for sake of simplicity  we analyse the two agents case, therefore one agent represents the source while the other the target of transfer.

We assume that agents explore at a different time the same or similar tasks (but performed in a different underlying environment). Thus, no knowledge mapping is required as both agents have the same set of sensors and actuators.
Algorithm~\ref{alg:source_agent}, describes the process of source agent that collects samples, represented as a tuple: $(s_t, a_t, r_t, s', u_{s_t})$, hence stores agent-environment interactions plus a confidence value computed over the visited state.
The latter value depicts state-related agent epistemic confidence at time $t$ of the visit. In detail, $u$ stands for uncertainty and therefore, higher is the value lower is the agent's confidence within that state.

\begin{algorithm}[tb]
	\caption{Process for source agent that stores agent-environment interactions.}
	\label{alg:source_agent}
	\begin{algorithmic}
		%\STATE {\bfseries Input:} data $x_i$, size $m$
		\STATE Initialize experience buffer $EB$
		\FOR{$e$ in episodes}
		\REPEAT
		\STATE observe state $s$
		\STATE take an action $a$
		\STATE observe action, reward $r$ and next state $s'$
		\IF{time to update policy}
		\STATE optimize policy for K epochs on collected interactions
		\STATE optimize RND on visited states
		\ENDIF
		\IF{$ep$ $>$ episode from saved experience}
		\STATE estimate agent confidence $u$ over $s$ through RND
		\STATE add ($s,a,r,s',u$) to buffer $EB$
		\ENDIF
		\UNTIL episode is complete or max steps is reached
		\ENDFOR
		\end{algorithmic}
\end{algorithm}

\begin{algorithm}[tb]
	\caption{Flow of target agent that samples interactions. }
	\label{alg:target_agent}
	\begin{algorithmic}
		\STATE {\bfseries Input:} experience buffer $EB$, threshold $t$, transfer budget $B$
		\STATE Initialize agent
		\STATE sample from $EB$ $B$ interactions where $u < t$
		\STATE train agent with experience sampled
		\FOR{$e$ in episodes}
		\REPEAT
		\STATE observe state $s$
		\STATE take an action $a$
		\STATE observe action, reward $r$ and next state $s'$
		\IF{time to update policy}
		\STATE optimize policy for K epochs on collected interactions
		\ENDIF
		\UNTIL episode is complete or max steps is reached
		\ENDFOR
		
	\end{algorithmic}
\end{algorithm}

Once the first agent has accomplished the training, buffer of collected experience is transferred to a receiving agent. The latter filters the received knowledge based on interaction's confidence $u$ and confidence threshold $t$  and draws a certain number of samples to pre-train its learning model as in Algorithm~\ref{alg:target_agent}.

This technique is applicable to RL algorithms regardless the underlying representation, e.g., tabular model or through a neural network. Our implementation uses policy-based deep RL algorithm called proximal policy optimization (PPO)~\cite{PPOagentOpenAI}. PPO is an on-policy algorithm that optimises a policy by gradient descent, meaning that the updated policy and the one used to produce samples are the same. Therefore, to not compromise the optimisation process, we limit the bootstrapping of external knowledge before the learning phase commence. While off-policy algorithms might be naturally more suitable for reusing experience, in this paper we based it on PPO, due to faster convergence rates %~\cite{someone} 
and its previous successes in ride-sharing application~\cite{castagna2020d2r2}.

%We opted for PPO because has already proven good performance in previous applications~\cite{castagna2020d, castagna2020demand} and enables agents to learn the task in short time.

\section{EVALUATION ENVIRONMENTS}\label{sec:simulation}

This section introduces the scenarios used to perform evaluation, and the parameters used in these scenarios.

\subsection{Predator-prey}
\begin{figure}[!h]
	\centering
	\includegraphics[width=0.6\columnwidth]{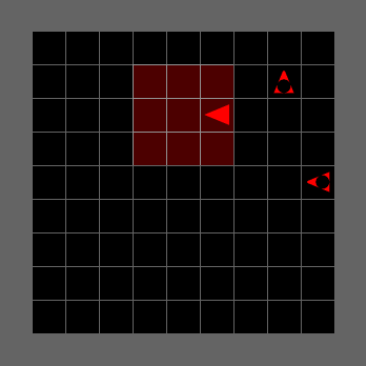}
	\caption{Instance of 9x9 grid predator prey environment with a single predator (filled triangle), and multiple preys (pierced triangles). Predator perceives the highlighted 3x3 grid in front of it and its goal is to catch the preys.}
	\label{fig:pp}
\end{figure}

Predator-prey environment is based on an existing version of a grid-world implementation~\cite{gym_minigrid}. Figure~\ref{fig:pp} shows a basic configuration of the environment. Task starts with a random allocation of predator and preys around the grid, while ends whether the predator, represented as a front-oriented filled triangle, catches all the same colour-based preys, depicted as pierced triangles. The environment is partially observable as the agent knows only the 3 x 3 grid in its field of vision, which in the figure is highlighted in front of it. To fulfil the goal, a predator is enabled to move forward, turn left or right, wait and perform a catch action. With the latter, a predator can catch a prey when it is in the consecutive cell.

Prey follows a random policy to escape from the predator, as follows: stay in the same position with 10\% of probability, turn left or right with 25\% chance each, and  move forward with 40\% probability.

The reward model is designed to encourage predator to catch the prey faster; the agent receives a living penalty according to the action taken, $-0.01$ for turn and step actions, $-0.25$ when staying still and finally $-0.5$ for a missed catch. However, whether the catch succeeds agent receives a positive reward of 1. The configuration used within this work is reported in Table~\ref{tab:pp_env_settings}.

For evaluation purpose, we use this benchmark scenario to study the impact of sharing agent-environment interactions from a pre-trained agent to a new "blank" agent. Source of transfer is an agent that performs well enough to accomplish the given task. During exploration, this agent labels interactions with its epistemic confidence over the visited state. Thus, confidence is estimated on the number of state visits updated to the timestep of the sampling.
Afterward, knowledge is filtered by a specified confidence threshold and transferred to the experience buffer of the new agent. The new agent samples a number of interactions from the buffer and pre-trains its learning model before exploring.

We study the impact of leveraging collected experience by varying threshold confidence level and number of interactions sampled. By doing so, we aim to gain insights into how is the second agent behaviour influenced by the amount of samples passed and by the quality of injected knowledge.

\subsection{Ride-Sharing-enabled Mobility on Demand Scenario}
For the real world case study we use the system presented in~\cite{castagna2020d2r2}. However, we port the implementation into a widely used traffic simulator of urban mobility, SUMO ~\cite{sumo2018}, to increase realistic traffic movement and conditions. 

Agent's goal is to maximise the amount of ride-request served. To serve all the requests, a fleet of 5-seater RS-enabled vehicles is dispatched. Control is distributed at vehicle level and therefore, each car is controlled by a single agent with no communication nor coordination with others.

\begin{figure}[!h]
	\centering
	\includegraphics[width=1\columnwidth]{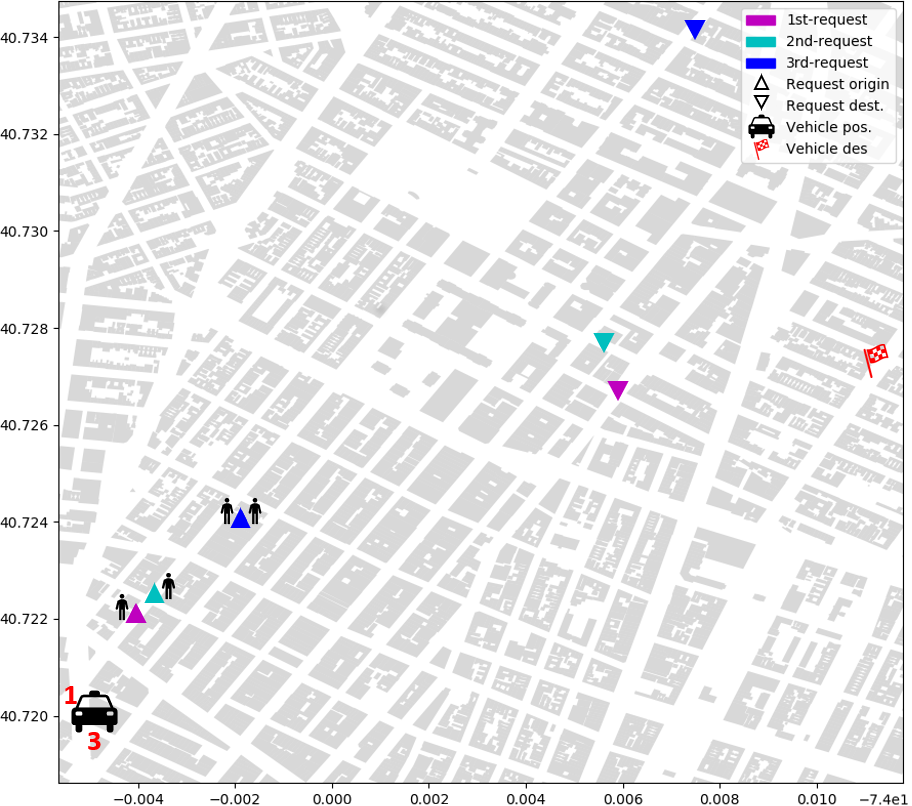}
	\caption{Representation of vehicle status with the three requests perceived in the neighbourhood.}
	\label{fig:rs}
\end{figure}

At each timestep, each RL vehicle agent knows its location, number of unoccupied seats and destination of the request that can be served the quickest, among those that it has assigned.
Furthermore, an agent receives information about the three closest requests in the neighbourhood, with an number of passengers that could fit on board and that could be picked up before request's expiration time.
A representation of state and perception is shown in Figure~\ref{fig:rs}. For each of these requests, agent knows request origin, request destination, number of passengers and minimum detour time from the current route to reach the pick-up point and destination point of the request.

After evaluating state and perception, an agent can take one of the 5 actions available, (1) \textit{being parked}, vehicle stays parked for a predefined amount of time, (2) \textit{drop-off}, vehicle drives towards its destination and finally, (3) \textit{pick-up}, where agent drives to pick-up point of the selected request. Note that agent can decide to pick one of the three available requests and as results, agent has three different pick-up options, resulting in total of 5 available actions.

Vehicle position is updated in real-time, however decision process is triggered on the conclusion of an action, i.e, vehicle arrives to pick-up or drop-off location. On top of that, a vehicle can evaluate to pick further requests while is  serving others. This evaluation step is allowed whenever the vehicle is further from arrival point and has at least a free seat. All vehicles within the fleet are synchronized under the same clock, therefore evaluation process is performed following a FIFO queue. Given the task's nature, an episode begins with the vehicle spawn and terminates when there are no further request to be served.

Reward scheme is designed to minimise the cumulated delay for each of the action taken by the agent.
%define delay
Therefore, is defined as the elapsed time from the agent's decision to the end of the action.
First, when an action is not possible to accomplish, agent receives a penalty of -1. Second, when agent decides of being parked, receives a penalty defined over the elapsed time: $\frac{-x}{x+delay}$, where $x$ is set to 5 whether agent is not serving any request, else to 1. Third, while dropping off, reward is defined as follow, $\exp{(\frac{1}{1+delay})}$.
Last, on pick-up reward given to an agent is defined as $\frac{x}{x+delay}$,where $x$ is fixed to 1 whether the request picked is the first, 2 otherwise. Reward model has been designed in that way to normalize delay into (0,1) interval.

%drop off defined as exp(1/1+delay)
%pick defined as x/ (x+delay) where x =1 if 1st request else 2
%park is -y/(y+delay) whre y=5 if no requests serving else 1
%not possible request is -1

%how is dataset organized:
To emulate a real-world scenario, we use ride-requests data from~\cite{samod2020}, where authors have aggregated NYC taxi trips from 50 consecutive Mondays between July 2015 and June 2016 in Manhattan zone~\cite{NYC_data}. Trips are then organized on a time-base schedule. Finally, we obtained 4 datasets representing peak times, morning, afternoon evening and night. Within this research, we use the morning peak slot (from 7 to 10 am) and the evening (from 6 to 9 pm).

%how we transfer?

We use these different datasets to evaluate the impact of sharing experience among same tasks but varying the underlying demand pattern. We let agents explore and collect interactions by serving requests from the morning set and afterwards we leverage part of the collected experience to pre-train agents that operate in the evening shift. In detail, 200 vehicles are dispatched during simulation to serve the demand.
%how evaluation is done?
To evaluate our work,  we use two demand loads and we execute the system in the following conditions, 1) train and test on morning peak; 2) train and test on evening peak; 3) train on morning peak and test on evening peak; lastly, 4) train on morning peak, transfer knowledge to a new agent that will then be further trained and tested on evening peak.
%#For the latter, we evaluate different threshold levels when filtering transferred experience.

%To compare our proposal against the baselines we study the undiscounted cumulated reward curve during learning. Furthermore, we evaluate performances according to MoD metrics such as, mean waiting time for requests, fleet mileage, average accomplished requests per vehicle.

%\subsection{Simulation Settings}

\section{EVALUATION RESULTS AND ANALYSIS}
\label{sec:results}

This section presents and discuss the results achieved through simulations in the two evaluation environments and is organized in two parts. First, we discuss the benefit of reusing experience in predator-prey scenario and second, in MoD scenario.

\begin{comment}
\subsection{Selecting RND parameters} 
We performed several empirical tests on the confidence estimator model, varying the number of predicted values within RND.
We limited the study on predator-prey environment as we expect the model to have comparable trend in other environments. Although, the threshold values are scenario dependent as the uncertainty is influenced by the size of state representation alongside the range that these features may have.

\begin{figure}[!h]
\centering
\includegraphics[width=.75\columnwidth]{assets/zoom_predator_uncertainty_analysis_10_runs.pdf}
\caption{Zoom of uncertainty trend in the first 1,000 episodes on predator-prey varying encoder dimension. The curve is obtained as average of 10 runs.}
\label{fig:RND_confidence_variation}
\end{figure}

As Figure~\ref{fig:RND_confidence_variation} shows, a low number of outputs to predict, i.e., 32 or 128, does not capture the changes across different states. On the other hand, the model seems too sensible when we have many neurons to predict, i.e., 4096. For our experiments, we, therefore, fixed the encoder size to 1024, which seems to capture similar states while differentiate when the agent encounters poor visited ones.

\end{comment}

\subsection{Predator-prey}
To evaluate the performance obtained within predator-prey scenario we studied agent's learning curve, by analysing the reward.

For the following set of experiments, Table~\ref{tab:pp_env_settings} summarises the configuration of the environment, while  Table~\ref{tab:sim_settings} reports the simulation parameters for the task. We set RND size to 1024 as we discovered to be a good trade off between performance and network size.

\begin{figure*}[h!]
	\centering
	\begin{subfigure}[t]{1\columnwidth}
		\centering
		\includegraphics[width=1\columnwidth]{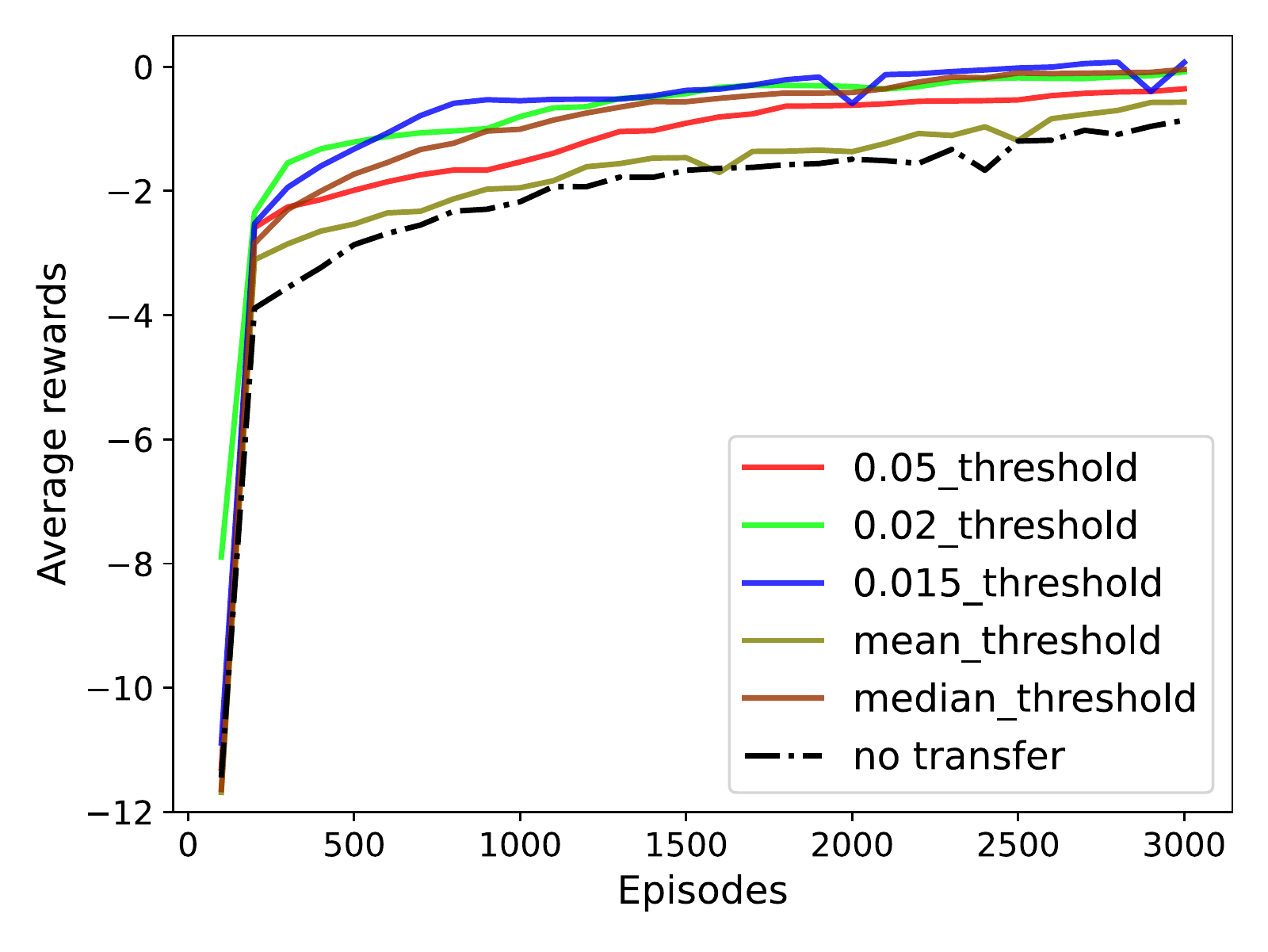}\caption{}
	\end{subfigure}~
	\begin{subfigure}[t]{1\columnwidth}
		\centering
		\includegraphics[width=1\columnwidth]{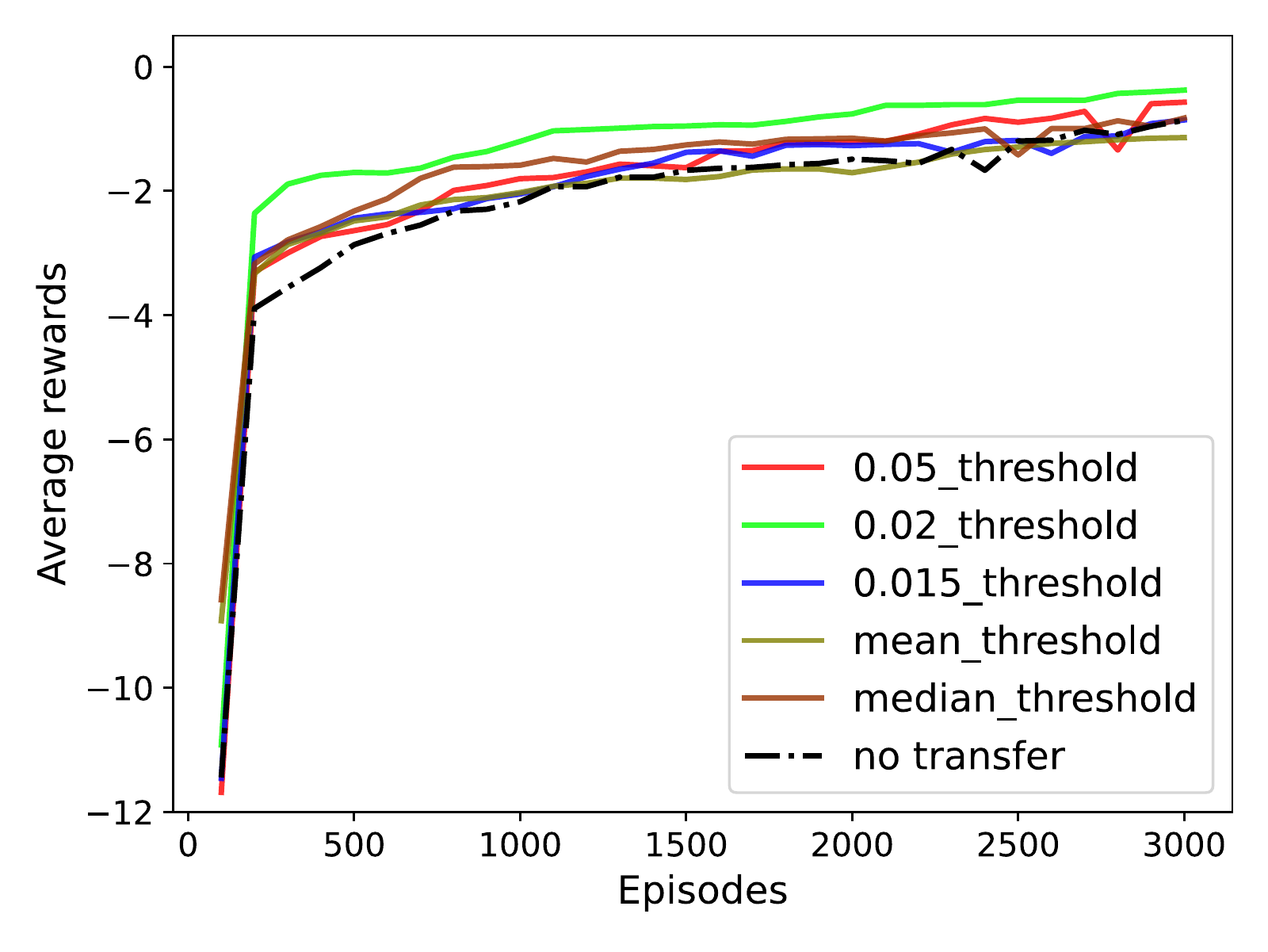}\caption{}
	\end{subfigure}
	\caption{Comparison of averaged reward with different number of interactions sampled by the transfer buffer, 5.000 \textit{(a)} and 10.000 \textit{(b)}. Results are obtained as average over 50 simulations in predator prey scenario enabling transfer learning.}
	\label{fig:res_pp_tb_size}
\end{figure*}

Within this work, we let the source agent to collect experience from the last 20\% episodes and we vary the number of drawn interactions from the transfer buffer (5,000 and 10,000) and confidence threshold for filtering interactions. Figure~\ref{fig:res_pp_tb_size} depicts the achieved results by varying both parameters. As threshold, we used few hardcoded task-related values empirically obtained (0.015, 0.02, and 0.05) and two others computed over the transfer buffer (mean and median confidence values).

\begin{figure*}[h!]
	\centering
	\begin{subfigure}[t]{1\columnwidth}
		\centering
		\includegraphics[width=1\columnwidth]{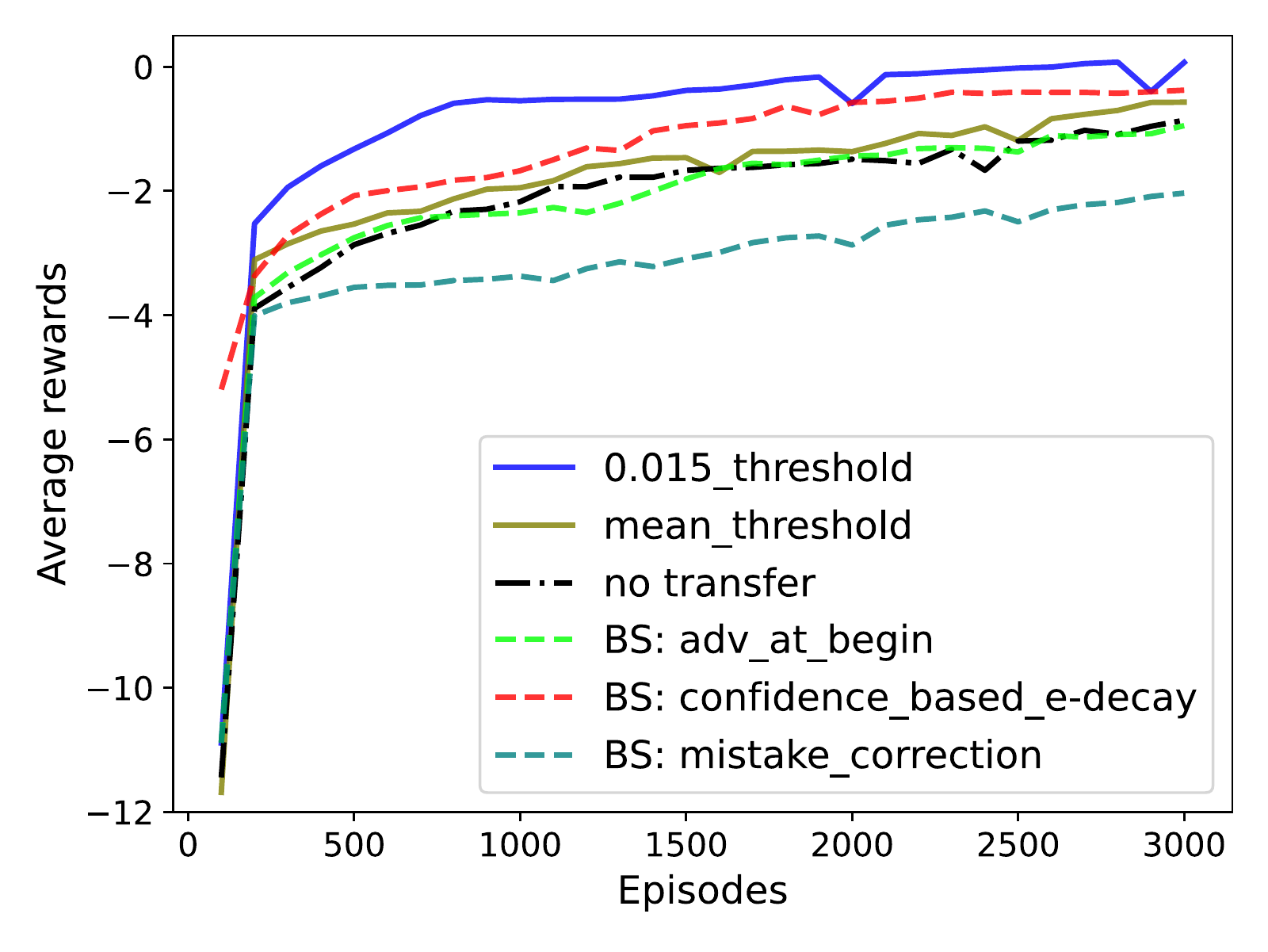}\caption{}
	\end{subfigure}~
	\begin{subfigure}[t]{1\columnwidth}
		\centering
		\includegraphics[width=1\columnwidth]{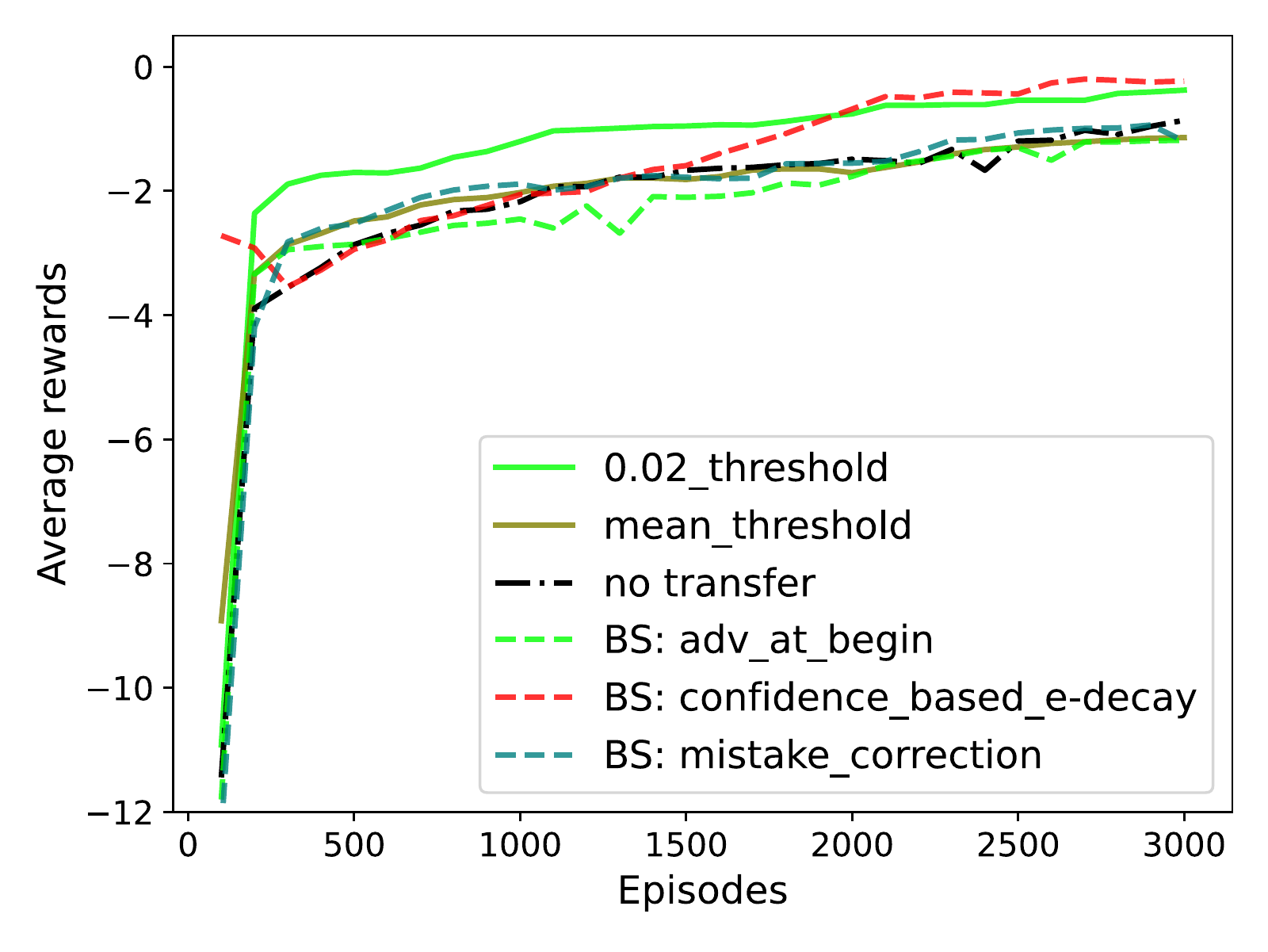}\caption{}
	\end{subfigure}
	\caption{Comparison of best and worst instance against proposed baselines with different number of interactions sampled by the transfer buffer, 5.000 \textit{(a)} and 10.000 \textit{(b)}. Results are obtained as average over 50 simulations in predator prey scenario enabling transfer learning.}
	\label{fig:res_baselines_pp}
\end{figure*}

In addition, we compare achieved results against four different baselines and we show the results through Figure~\ref{fig:res_baselines_pp}.
As first baseline, we propose a no-transfer agent, i.e., an agent that addresses the task for the first time with no external support.
Other baselines included an additional agent that can support the learning phase by providing advice when needed.
As transfer framework for these baselines, we opted for the teacher-student paradigm~\cite{torrey2013teaching}, where student is represented by the learning agent while teacher is the additional agent that already accomplished the task obtaining good performance.
Among the transfer-enabled baselines we can distinguish three cases:  first \textit{advice at beginning}, as we noticed that the initial episodes are the hardest for an agent to exhibit good behaviour, second \textit{mistake correction}, where teacher supervises student by providing advice when the latter misbehaves according to teacher's expectation and last, \textit{confidence based $\epsilon-$decay}~\cite{Norouzi2020experience}, where student asks for advice following an  $\epsilon-$decay probability when it has lower confidence than teacher. In all teacher-supported baselines, as is the standard in teacher-student TL related work, we set a maximum budget for exchanging advice in order to match the number of sampled interactions.

Figure~\ref{fig:res_pp_tb_size} shows the results achieved by varying filter threshold on receiving side organized by the amount of sampled interactions, 5,000 (\textit{a}) and 10,000 (\textit{b}). Impact of threshold is mild compared to transfer buffer size; contrary to what we intuitively expected, using less samples (5,000) has shown a greater improvement of the performance against a no transfer enabled agent.
Regardless of the transfer settings, agent performance is overall better compared to a no transfer agent. In a few of the cases we observe a jumpstart, i.e., an improved performance since the very beginning. However, we cannot do a linear comparison between transfer and no-transfer results because the former exploits external knowledge resulting in a few additional episodes. Nevertheless, most of transfer-enabled instances are able to outperform a no transfer agent since the beginning as we can easily notice from Figure~\ref{fig:res_pp_tb_size}a.

Tuning the right threshold could be very expensive in term of time as it is task-dependent and might not results in the expected outcome (i.e., resulting in a negative transfer). Nevertheless, Figure~\ref{fig:res_baselines_pp} shows a significant gap between our proposal and the evaluated baselines.
In particular, we can notice from Figure~\ref{fig:res_baselines_pp}\textit{a} that when the transfer budget is low, our proposal stands out from the others and enables the agent to reach very soon good performance that are kept over time.

%Even if could be very expensive to tune the right threshold in order to optimize the results

\begin{table}[!htb]
	\small
	\renewcommand*{\arraystretch}{1.15}
		\caption{\label{tab:pp_env_settings} Parameters in predator-prey environment.}
		\centering
		\begin{threeparttable}
			\begin{tabularx}{0.9\columnwidth}
				{|>{\centering\arraybackslash}c|>{\centering\arraybackslash}c|}
				\cline{1-2}
				{Parameter}&{Value}\\
				\cline{1-2}				
				%start content
				\multicolumn{1}{|c|}{\textit{grid size}}
				& 9\\
				\cline{1-2}
				\multicolumn{1}{|c|}{\textit{\#teams}}
				& 1\\
				\cline{1-2}
				\multicolumn{1}{|c|}{\textit{\#predators}}
				& 1\\
				\cline{1-2}
				\multicolumn{1}{|c|}{\textit{\#preys}}
				& 2\\
				\cline{1-2}
				\multicolumn{1}{|c|}{\textit{grid-view size}}
				& 3\\\cline{1-2}
				\multicolumn{2}{|c|}{{rewards}}\\
				\cline{1-2}
				\multicolumn{1}{|c|}{\textit{successful catch}}
				& 1\\
				\cline{1-2}
				\multicolumn{1}{|c|}{\textit{missed catch}} %no agent that tries to catch every time
				& -.5\\
				\cline{1-2}
				\multicolumn{1}{|c|}{\textit{hold position}} %no lazy agent waiting for prey to pass by
				& -.25\\
				\cline{1-2}
				\multicolumn{1}{|c|}{\textit{turn or step penalty}} % agent no waste of time
				& -.01\\
				
				%\cline{1-2}
				%\multicolumn{2}{|c|}{\textbf{prey actions distribution}}\\
				%\cline{1-2}
				%\multicolumn{1}{|c|}{\textit{step forward}}
				%& .4\\
				%\cline{1-2}
				%\multicolumn{1}{|c|}{\textit{turn (left,right)}}
				%& .25\\
				%\cline{1-2}
				%\multicolumn{1}{|c|}{\textit{hold}}
				%& .1\\
				\cline{1-2}			
			\end{tabularx}
		\end{threeparttable}
\end{table}

\subsection{Mobility on Demand}

For this scenario we adopted a MoD previously evaluated against multiple baselines with and without ride-sharing and rebalancing~\cite{castagna2020d2r2}, therefore in this work we solely focus on evaluating the impact of TL. Table~\ref{tab:mod_env_settings} presents the environment related settings while, Table~\ref{tab:sim_settings} reports the learning model configuration.

\begin{table}[!htb]
	\small
	\renewcommand*{\arraystretch}{1.15}
	\caption{\label{tab:mod_env_settings} Parameters in Mobility-on-Demand scenario.}
	\centering
	\begin{threeparttable}
		\begin{tabularx}{0.9\columnwidth}
			{|>{\centering\arraybackslash}c|>{\centering\arraybackslash}c|}
			\cline{1-2}
			{Parameter}&{Value}\\
			\cline{1-2}				
			%start content
			\multicolumn{1}{|c|}{\textit{fleet size}}
			& 200\\
			\cline{1-2}
			\multicolumn{1}{|c|}{\textit{\#seats}}
			& 5\\
			\cline{1-2}
			\multicolumn{1}{|c|}{\textit{\#available requests 7-10am}}
			& 9663\\
			\cline{1-2}
			\multicolumn{1}{|c|}{\textit{\#available requests 6-9pm}}
			& 9662\\
			\cline{1-2}
			\multicolumn{1}{|c|}{\textit{training rounds}}
			& 6\\\cline{1-2}
			\multicolumn{1}{|c|}{\textit{training vehicle per round}}
			& 10\\\cline{1-2}
			\cline{1-2}			
		\end{tabularx}
	\end{threeparttable}
\end{table}

%Reward model is a bit more sophisticated compared to the other scenario and can be described as follow. When the vehicle is parked, receives a penalty of $\frac{-x}{x+park time}$, where $x$ is 1  if the vehicle has not any passengers onboard, otherwise is 5. When the vehicle arrives to passengers' destination reward is $\exp^{\frac{1}{1+trip time}}$ where $trip time$ represent the lapse of time since the onboarding of passengers. With pick-up action agent receives a reward of $\frac{x}{x+time\ for\ pick}$, where $x$ is 1 if the vehicle before picking up was empty and 2 otherwise while $time\ for\ pick$ represents elapsed time since the request assignment to vehicle.

\begin{table}[!htb]
	\small
	\renewcommand*{\arraystretch}{1.15}
	\centering
	\caption{\label{tab:sim_settings} Simulation settings for predator-prey (PP) and Mobility-on-Demand (MoD) scenarios.}
	\begin{threeparttable}
		\begin{tabularx}{0.9\columnwidth}
			{|>{\centering\arraybackslash}c|>{\centering\arraybackslash}c|>{\centering\arraybackslash}c|}
			\cline{1-3}
			{Parameter}&{PP} & {MoD}\\
			\cline{1-3}				
			%start content
			\multicolumn{1}{|c|}{\textit{\# episodes}}
			& 3000 & N/A\\
			\cline{1-3}
			\multicolumn{1}{|c|}{\textit{max \#steps}}
			& 500 & N/A\\
			\cline{1-3}
			\multicolumn{1}{|c|}{\textit{\#hidden layer(s)}}
			& 1 & 4\\
			\cline{1-3}
			\multicolumn{1}{|c|}{\textit{size hidden layer(s)}}
			& 64 &128\\
			\cline{1-3}
			\multicolumn{1}{|c|}{\textit{gamma}} %discount factor
			& .99 &.999\\\cline{1-3}
			\multicolumn{1}{|c|}{\textit{learning rate}} %lr
			& .001 &1e-4\\\cline{1-3}
			\multicolumn{1}{|c|}{\textit{update\_iter}} %update policy every n timesteps
			& 500 & 32\\\cline{1-3}
			\multicolumn{1}{|c|}{\textit{epochs}} %update policy for k_epochs
			& 10 & 10\\\cline{1-3}
			\multicolumn{1}{|c|}{\textit{$\epsilon-$clip}} %clip value for ppo
			& .2 & .2\\\cline{1-3}
			\multicolumn{1}{|c|}{\textit{RND size}}
			& 1024 & 1024\\%lie is 1000
			\cline{1-3}			
		\end{tabularx}
	\end{threeparttable}
\end{table}

 We rely on common metrics used to evaluate performances of MoD system. On passenger side, we evaluate waiting time, defined as time elapsed from moment that request is being made until the passenger pick-up by the ride-sharing vehicle. Furthermore, as Table~\ref{tab:mod_results} reports, we take into account detour ratio ($\overline{D_r}$) that captures the additional distance driven by the vehicle with the passenger over the expected distance without ride-sharing enabled.
At the overall fleet level we evaluate percentage of requests being served in ride-sharing($\%RS$) and demand satisfaction rate ($\%\ Served\ Reqs.$).
Finally, on vehicle level we evaluate the passenger distribution across the available vehicles, reporting the variance of passengers distribution ($\mathbf{\sigma}\ pass$) in Table~\ref{tab:mod_results}, alongside the average travelled distance covered by vehicles during the whole simulation ($\overline{d_t}$).

We compare our TL-enabled MoD, \textit{TL-enabled test 6-9pm} against two baselines: 
\begin{enumerate}
\item \textit{Train 7-10am test 6-9pm}, where we transfer the learnt model from an agent trained on the morning demand set (7-10am) to evening set (6-9pm)
\item \textit{Train \& test 6-9pm} where an agent learns over the evening set and is evaluated over the same dataset. 
\end{enumerate}
We also present the results achieved on the morning peak by agent trained on the same set of requests~(\textit{Train \& test 7-10am}). The latter provides experience that we have used within our proposal where a novel agent, following Algorithm~\ref{alg:target_agent}, preprocesses received knowledge and then performs a training on evening demand set (6-9pm).

The details of how we conducted experience in our TL-based scenario \textit{TL-enabled test 6-9pm} are as follows. To collect enough samples, agent training in \textit{Train \& test 7-10am} required a larger set of requests. Hence we used 30,000 requests maintaining the scheme of 6 rounds with 10 vehicles, forming a 60 episodes training. In detail, each vehicle selects requests to be served from the pool of requests that previous vehicles could not accomplish. Once a round is completed, request set is restored and a new cycle can begin. All vehicles contribute to the same model, which is later replicated to all the 200 vehicles operating in morning and evening shifts.

 Agent was enabled to store knowledge only over the final 40\% episodes and it collected approximately 14,000 interactions. Uncertainty interval for the collect interactions was really broad, 17 - 500,000, with average being ~50,000 and with approximately 6,000 interactions having a state with no requests in agent neighbourhood and an associated uncertainty lower than 450.

Initially, we performed the tests with multiple thresholds applied to uncertainty level of interactions reused. First, we provided to target agent the whole set (approximately 14,000 interactions) without applying any filter. Second, we discarded all interactions composed by a state with no requests in proximity of the vehicle and from those we sampled 5,000 interactions. Third, we set the threshold to 50,000, utilizing only samples with uncertainty under that number, resulting in around 10,000 available interactions.

However, all of these initial tests failed to run to completion because they overloaded the SUMO engine with requests leading to a crash. Nevertheless, the early stages of the experiments were useful in order to tune the threshold. For the final experiment, results of which we present here we halve the available interactions for target agent by setting the threshold to 10,000 taking all the interactions with lower uncertainty. Agent randomly samples 5,000 interactions and pre-train its learning model. Finally, it performs a 6 rounds with 10 vehicles training.

\begin{figure*}[h!]
	\centering
	\begin{subfigure}[t]{2\columnwidth}
		\centering
		\includegraphics[width=1\columnwidth]{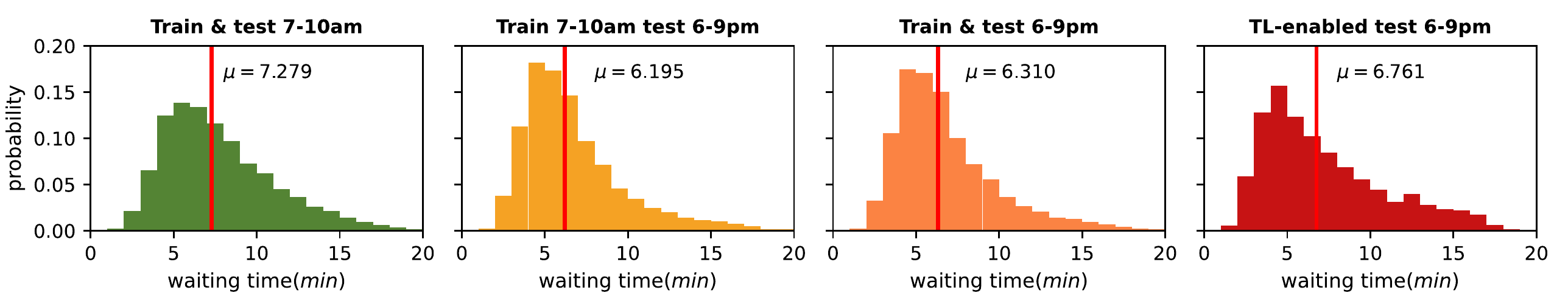}\caption{Waiting time for request}
	\end{subfigure}\\
	\begin{subfigure}[t]{2\columnwidth}
		\centering
		\includegraphics[width=1\columnwidth]{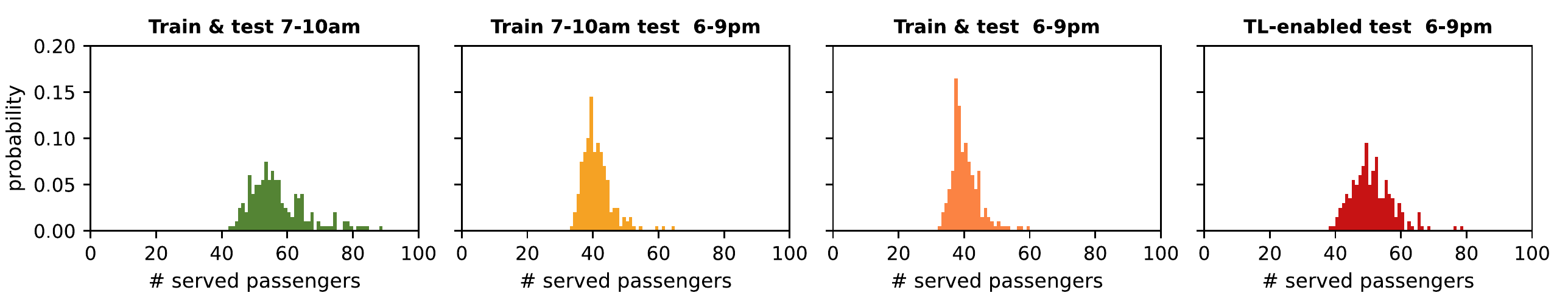}\caption{Number of passengers served per vehicle}
	\end{subfigure}\\
	\begin{subfigure}[t]{2\columnwidth}
		\centering
		\includegraphics[width=1\columnwidth]{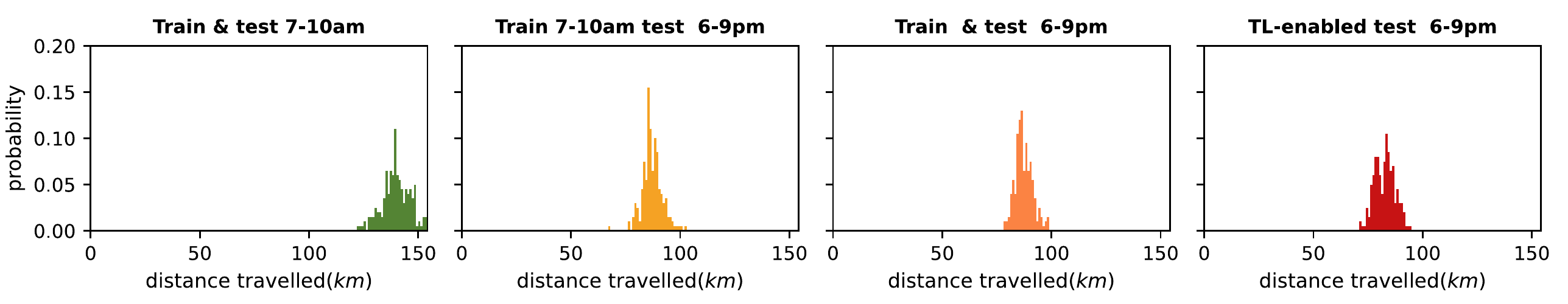}\caption{Vehicles mileage}
	\end{subfigure}
	\caption{Comparison of simulation results for the analysed scenarios over the following metrics: ($a$) waiting time, ($b$) passengers distribution per vehicle and ($c$) distance travelled by vehicle. The $y$-axis represents the probability that a request for ($a$) or vehicle in ($b$, $c$) assumes a certain value. Note that results are obtained from a single simulation run.}
	\label{fig:res_mod}
\end{figure*}

We present the simulation results in Table~\ref{tab:mod_results} and Figure~\ref{fig:res_mod}.  We show the waiting time for requests in ($a$), the distribution of passengers served per vehicle in ($b$) and the travelled distance per vehicle in ($c$).

\begin{table*}
	%\small % Try this, I'm not sure it will work, but the table can be made smaller %it actually makes it bigger
	\renewcommand*{\arraystretch}{1}
	\setlength{\extrarowheight}{1pt}
	\centering
	\caption{\label{tab:mod_results}{Performance metrics across all 5 evaluated approaches in ride-sharing scenario}.}
	\begin{tabularx} {1.7703\columnwidth}{|>{\centering\arraybackslash}p{4cm}|>{\centering\arraybackslash}p{2.75cm}|>{\centering\arraybackslash}p{.9cm}|>{\centering\arraybackslash}p{1.2cm}|>{\centering\arraybackslash}p{1.1cm}|>{\centering\arraybackslash}p{0.7cm}|}
		
		\hline
		\multirow{1}{*}{{Scenarios}} &{\%Served requests}&\multirow{1}{*}{{\%RS}}& ${\sigma}$  {pass} & ${\overline{d_t}}$ {\textit{(km)}} &\multirow{1}{*}{${\overline{D_r}}$}\\ 
		%&{Reqs.}& &{pass} & {\textit{(km)}}&\\ 
		\hline
		Train \& test 7-10am &  \multirow{1}{*}{93} & \multirow{1}{*}{93} &\multirow{1}{*}{76.61}  &\multirow{1}{*}{140} &\multirow{1}{*}{9.4} \\
		\hline
		Train 7-10am test 6-9pm &  \multirow{1}{*}{77} & \multirow{1}{*}{99}  & \multirow{1}{*}{21.65} & \multirow{1}{*}{87}&\multirow{1}{*}{9}\\
		\hline
		Train\&test 6-9pm & \multirow{1}{*}{76} &  \multirow{1}{*}{99}  & \multirow{1}{*}{19.35} & \multirow{1}{*}{87}&\multirow{1}{*}{9.11}\\
		\hline
		{TL-enabled test 6-9pm}  &\multirow{1}{*}{{79}}  &\multirow{1}{*}{{93}} & \multirow{1}{*}{{41.12}} & \multirow{1}{*}{{82}}&\multirow{1}{*}{{5.23}}\\
		\hline
	\end{tabularx}
\end{table*}

Note that green scenario, (\textit{Train \& test 7-10am}), reports a longer distance covered by vehicles due to a greater number of requests satisfied. 
Our scenario \textit{TL-enabled test 6-9pm}, enable the system to cover further requests when compared to baselines on same demand set. As Table~\ref{tab:mod_results} reports, it enables the fleet to cover 79\% of demand while \textit{Train 7-10 test 6-9pm} and \textit{Train \& test 6-9pm} stop at 77\% and 76\%. Furthermore, despite the few additional requests served,  average vehicles mileage is lowered by 5 km per vehicle (82 km) against (87km). %distance covered by vehicles during service is lowered by 5 km per vehicle (82 km) against (87km).
Although, passengers distribution is unbalanced compared to baselines, as variance is almost doubled (41.12) against (21.65 and 19.35)

%Although, passengers distribution variance over the fleet is higher (41.12) compared to baselines (21.65 and 19.35), resulting in an unbalanced dist few vehicles that are more performant in our proposed scenario. 
Eventually, while maintaining a fair RS rate (93) compared to baselines (99), our scenario almost halves the average request detour ratio (5.23) compared to baselines (9 and 9.11), resulting in less kilometres travelled onboard for passengers before reaching their destination.
%rario %rs decreasesd

When leveraging experience across equal-defined task as within our predator-prey simulations, target agent is enabled to outperform a no-transfer agent and baselines. In detail, TL-enabled agent overcomes no-transfer enabled agent performance in less than a half episodes.

On the other hand, when applying TL to a more complicated scenario, as our real-world MoD simulator, it demands a not trivial evaluation over transfer parameters. However, in our experiment, agents are able to satisfy a greater number of passengers while reducing travelled distance and improving rider experience. In fact, detour ratio, defined as distance travelled by passengers onboard of MoD vehicle against expected travelled distance on a solo trip, is sharply decreased when compared against other scenarios.

\section{CONCLUSION AND FUTURE WORK}
\label{sec:fwork}

In this paper we presented an experience-sharing (transfer learning) strategy to leverage collected experience across same tasks executed in different environment dynamics. A first agent~(source), collects samples throughout exploration while learning a task, that are then transferred to a second agent~(target). The latter, preprocesses its learning model by sampling a certain amount of interactions from receiving buffer. We have analysed the performance varying transfer parameters in two scenarios, predator-prey, where target and source agent address the very same task, and a mobility on demand scenario where source and target are evaluated on a same task but under different underlying ride request demand.

%conclusions
We generally noticed that within the predator-prey scenario, by enabling TL,  agent which receives transferred data is able to outperform a standard agent which learns from scratch. %By further tuning the TL parameters, such as uncertainty threshold level and transfer batch size, TL agent is able to outperform a standard agent.

These improvements are more difficult to notice in a simulated real world environment since we do not track the reward achieved by the agents but we rely on domain-specific performance measures dependent on our application area. However, we can still notice an improvement in the ride-requests satisfied and in the distance travelled by the vehicles. When we compare TL-enabled agent against agent learning from scratch on the same set of demand, we can notice that ride-sharing ratio is slightly decreased in favour of a lower detour ratio. As a result, passengers lower the time spent onboard by reaching faster their final destination.

This paper presents a first step in a wider work on multi-agent experience sharing in real-time. In this research, we evaluated the impact and the feasibility of defining parameters and threshold in experience reusing with fixed roles of information sender and receiver.

As a next step, we plan to evaluate "importance" of the states for the target agent, in order to reduce transfers in less important states to save communication budget if needed. For example, we have observed that a source agent has high confidence in its knowledge about the state in which there are no requests to serve, but since there is less potential for "mistake" actions when there are no requests, transferring knowledge in this situation might not be crucial.
%Before moving to real-time experience sharing, where roles of sender and receiver are interchangeable, we must carry out further evaluation tests on threshold definition in order to study the relation between a "kind of interaction",
%TODO alberto: i dont get what "kind of interaction" is and what do you mean by relation between it, and what else? between different kinds of interactions? what are interactions? the example below doesnt really help either, so maybe reword both xplaination and the example
Afterward, we plan to bootstrap knowledge across scenarios with different road structure to further test the reliability of TL applied to simulation of real-world case and develop the full dynamic online multi-agent TL system.

%When leveraging experience across equal-defined task as within our predator-prey simulations, target agent is enabled to outperform a no-transfer agent and baselines. In detail, TL-enabled agent overcomes no-transfer enabled agent performance in less than a half episodes.

% On the other hand, when applying transfer learning to a more complicated scenario, as our real-world MoD simulator, it demands a not trivial evaluation over transfer parameters. However, in our experiment, agents are able to satisfy a greater number of passengers while reducing travelled distance and improving rider experience. In fact, detour ratio, defined as distance travelled by passengers onboard of MoD vehicle against expected travelled distance on a solo trip, is sharply decreased when compared against other scenarios.

\section*{ACKNOWLEDGEMENTS}{This work was sponsored, in part, by the Science Foundation Ireland under Grant No. 18/CRT/6223 (Centre for Research Training in Artificial Intelligence), and Grant No. 16/SP/3804 (Enable).}
\bibliography{ref}
\bibliographystyle{apalike}

\end{document}